\documentclass[runningheads]{llncs}
\usepackage[T1]{fontenc}
\usepackage{graphicx}
\usepackage{multirow}
\usepackage{booktabs}
\usepackage{subcaption}
\usepackage{float}   
\usepackage{adjustbox} 
\usepackage{wrapfig}
\usepackage{caption} 
\usepackage{amssymb}
\usepackage{siunitx}
\usepackage{amsmath}
\begin{document}
\title{PINN-Cast: Exploring the Role of Continuous-Depth NODE in Transformers and Physics Informed Loss as Soft Physical Constraints in Short-term Weather Forecasting}
%
\titlerunning{PINN-Cast}

%
\author{Hira Saleem\inst{1}\orcidID{0009-0000-9046-0446} \and
Flora Salim\inst{1}\orcidID{0000-0002-1237-1664} \and
Cormac Purcell\inst{1}\orcidID{0000-0002-7491-7386}}
\authorrunning{H. Saleem et al.}
%
%
\institute{University of New South Wales, Sydney, NSW, Australia 
\\
\email{\{h.saleem,flora.salim,cormac.purcell\}@unsw.edu.au}}
\maketitle              
\begin{abstract}
Operational weather prediction has long relied on physics-based numerical weather prediction (NWP), whose accuracy comes at the cost of substantial compute and complex simulation workflows. Recent transformer-based forecasters offer efficient data-driven alternatives, however transformers are physics-agnostic models. Additionally, standard transformer encoders evolve representations through discrete layer updates that may be less suited to modeling smooth latent dynamics. In this work, we propose a continuous-depth transformer encoder for weather forecasting that integrates Neural Ordinary Differential Equation (Neural ODE) dynamics within each encoder block. Specifically, we replace discrete residual updates with ODE-based updates solved using adaptive numerical integration. We also introduce a two-branch attention module that combines conventional patch-wise self-attention with an auxiliary branch that applies a derivative operator to attention logits, providing an additional change-sensitive interaction signal. To further align forecasts with governing principles, we propose a customized physics-informed training objective that enforces physical consistency as a soft constraint. We evaluate the proposed method against a standard discrete transformer baseline and an existing continuous-time Neural ODE forecasting variant, demonstrating the importance of PINN-Cast in short term weather forecasting. 

\end{abstract}

\section{Introduction}
Weather forecasting can be viewed as a continuously evolving physical process governed by time-dependent partial differential equations (PDEs) that describe the spatiotemporal evolution of key atmospheric variables (e.g., temperature, pressure, humidity, and wind). In operational settings, these governing equations are solved numerically via large-scale simulation pipelines (numerical weather prediction, NWP), which require sophisticated discretizations, high-performance computing (HPC) resources, and careful management of computational cost \cite{andersson2022medium,bauer2020ecmwf,palmer2005representing}. 

While such full-fidelity NWP systems remain the foundation of operational forecasting, they are also complex to develop, expensive to run at scale, and difficult to iterate on rapidly when exploring new modeling ideas. Motivated by the broader goal of advancing computationally efficient scientific modeling, this work is positioned as an exploratory study which investigates whether (i) continuous-depth Neural ODE (NODE) dynamics inside transformer blocks and (ii) physics-informed soft constraints can improve learning stability and physical plausibility in data-driven forecasting at moderate resolution.

Recent advances in artificial intelligence (AI), particularly deep learning and transformer-based architectures particularly Vision Transformer (ViT) \cite{dosovitskiy2020image}, have enabled data-driven weather forecasting models that achieve strong predictive accuracy and high throughput \cite{bi2022pangu,nguyen2023climax,nguyen2023scaling}. However, typical transformer encoders update hidden representations through a fixed number of discrete residual steps, which may be less suited to representing smoothly evolving dynamics. This motivates hybrid architectures that integrate continuous-depth dynamics into attention-based models.

To advance physics-aware forecasting within a computationally efficient transformer framework, we propose \textbf{PINN-Cast}, a continuous-depth transformer encoder for weather forecasting by integrating Neural Ordinary Differential Equation (NODE) \cite{chen2018neural} updates inside each transformer block. Specifically, we represent the hidden state evolution as an initial value problem, where
$
\frac{dh(t)}{dt} = f_{\theta}(h(t), t),
$
and $f_{\theta}$ is parameterized by a neural network. In our implementation, $f_{\theta}$ is realized as a lightweight multilayer perceptron. These NODE \cite{chen2018neural} updates replace the conventional discrete residual transformations applied after attention and after the feed-forward network, yielding continuous-depth feature evolution within the encoder.

In addition, we propose a two-branch attention mechanism that augments standard patch-wise self-attention with an auxiliary derivative-based branch. The derivative branch computes finite differences of attention logits providing a complementary change-sensitive interaction signal. Finally, we incorporate physical constraints through soft penalties via a customized physics-informed loss derived from fundamental relationships associated with thermodynamics and kinetic energy. This objective discourages violations of established physical couplings between temperature and wind evolution, guiding learning toward forecasts that are not only data-consistent but also physically plausible. In summary, our contributions are as follows:

\begin{enumerate}
    \item \textbf{Two-branch attention with derivative-based interaction:} We propose an attention module that fuses patch-wise self-attention with an auxiliary branch that applies a finite-difference operator to attention logits introducing a change-sensitive interaction signal.
    \item \textbf{Continuous-depth transformer encoder via NODEs:} We introduce NODE layers within the transformer encoder to replace discrete residual updates with continuous-depth latent dynamics, yielding smoother and more stable representation evolution.
    \item \textbf{Physics-informed soft constraints:} We design a customized physics-informed loss based on thermodynamic and kinetic-energy relationships to penalize physically inconsistent behavior during training.
    \item \textbf{Evaluation:} We evaluate PINN-Cast on WeatherBench at $5.625^\circ$ resolution against (i) a discrete-depth transformer baseline and (ii) a continuous-time Neural ODE-based forecasting variant, demonstrating the impact of ODE-based encoder updates and physical consistency on forecast accuracy.
\end{enumerate}

\section{Related Work}
Currently, operational weather forecasting systems (IFS) \cite{documentation2023part} approximate atmospheric evolution by numerically discretizing and integrating the governing equations, while representing unresolved processes through physical parameterizations. While highly accurate, the forecast skill is bounded by initial-condition uncertainty, model error/parameterizations and high computational costs. 

Meanwhile, recent data-driven forecasting models learn forecast operators directly from data and deliver competitive skill with substantially lower inference cost, motivating hybrid and physics-aware approaches alongside traditional NWP. FourCastNet \cite{pathak2022fourcastnet,kurth2023fourcastnet} introduced operator-learning for global forecasting showing that Fourier based models can approximate atmospheric evolution with very fast inference. On the other hand, GraphCast \cite{lam2023learning} achieved state-of-the-art medium-range performance using multi-mesh graph neural networks.
Utilizing transformer's long range dependency, weather forecasting models such as Pangu-Weather \cite{bi2023accurate} demonstrated strong medium-range skill with a 3D Earth-specific transformer. Building on the need for broader generalization across variables and tasks, ClimaX \cite{nguyen2023climax} reframed forecasting as a foundation-model problem enabling flexible multi-variable, multi-task transfer. Subsequently, FengWu \cite{chen2023fengwu} and FuXi \cite{chen2023fuxi} advanced long-lead forecasting via large-scale architectures and multi-stage/cascaded strategies to control error growth.
Stormer \cite{nguyen2023scaling} further showed scalable transformer forecasters with improved robustness across lead times through training objectives for variable dynamics. ArchesWeather \cite{couairon2024archesweather} revisited efficient attention and architectural priors to improve practicality without sacrificing accuracy. 

Complementing these discrete space–time predictors, ClimODE \cite{verma2023climode} introduced a continuous-time, Neural ODE-based formulation to better reflect atmospheric transport dynamics (e.g., advection) and move forecasting toward smoother latent evolution, highlighting the promise of continuous-depth modeling even when peak accuracy still lags the largest discrete forecasters. Beyond purely data-driven evolution, NeuralGCM \cite{kochkov2023neural} moved further toward physics-consistent learning by coupling neural parameterizations with a differentiable dynamical core without sacrificing skill.

While the above models demonstrate impressive accuracy and scalability, a large subset of state-of-the-art neural forecasters still advance time through discrete rollouts and evolve internal features via discrete layer-wise residual updates, and in many purely data-driven designs physical consistency is learned only implicitly from data rather than being explicitly encouraged. In contrast, continuous-time and physics-aware directions such as ClimODE \cite{verma2023climode} and NeuralGCM \cite{kochkov2023neural} highlight the value of incorporating smoother dynamics and physical structure, but they also motivate the need for architectures that can retain transformer-level expressivity while integrating continuous-depth evolution and lightweight physics regularization. Addressing this gap, our work replaces discrete residual updates with Neural ODE dynamics solved via adaptive integration, augments standard attention with a change-sensitive two-branch mechanism (including a derivative-operator signal on attention logits), and introduces a custom physics-informed training objective as a soft constraint, together aiming to improve short-term forecast fidelity while better aligning learned dynamics with governing principles.

\section{Methodology}
\subsection{Notation}
Throughout this section, we use the following notation. $B$ denotes the batch size, $H$ and $W$ denotes the Latitude-Longitude grid. $V$ is the number of atmospheric variables. $N$ the number of patch tokens, $C$ the embedding dimension, $N_h$ the number of attention heads, and $d_h = C/N_h$ the per-head dimension. $W_{\mathrm{qkv}} \in \mathbb{R}^{C \times 3C}$ is a learned linear projection that produces queries, keys, and values jointly.

\subsection{Problem Formulation}
Given an input multivariate weather field $X_t \in \mathbb{R}^{V \times H \times W}$ at time $t$. The goal is to learn a forecaster $f_\theta$ that predicts the future state at a lead time $\Delta$ (e.g., $\Delta$ hours). All forecasts are produced via a single forward pass. The model directly predicts the state at each target lead time $\Delta$.
\begin{equation}
\hat{X}_{t+\Delta} = f_\theta\!\left(X_t, \Delta\right), 
\qquad 
\hat{X}_{t+\Delta} \in \mathbb{R}^{V \times H \times W}.
\end{equation}

\subsection{Two-Branch Attention with Neural ODE Integration:}
\begin{figure}[ht]
\begin{center}
\centerline{\includegraphics[width=\textwidth]{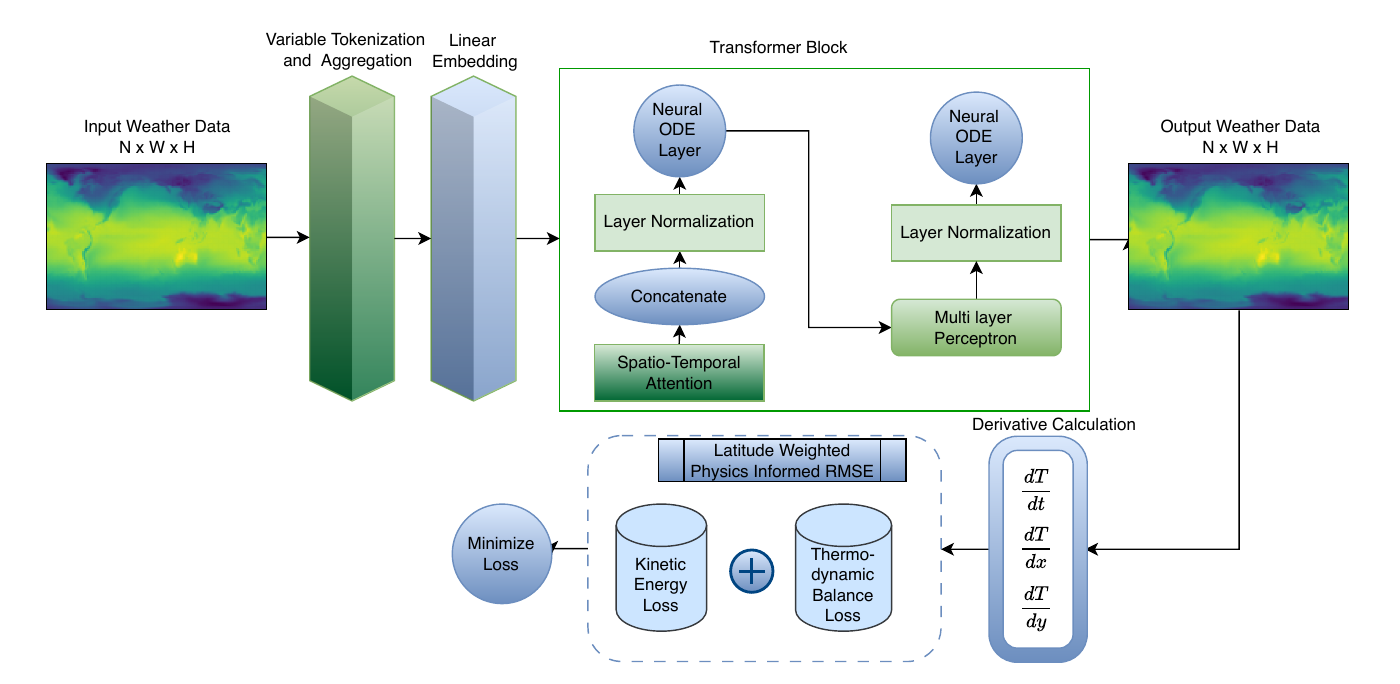}}
\caption{Overall prediction pipeline. The model receives spatiotemporal weather fields as input, processes them through transformer encoder blocks containing two-branch attention and Neural ODE updates, and produces the forecast $\hat{X}_{t+\Delta}$. During training (dashed box), the physics-informed loss supervises the entire forward pass by combining latitude-weighted MSE with kinetic energy and advection consistency penalties}
\label{fig:architecture}
\end{center}
\end{figure}

We extend the ClimaX variable-tokenization framework \cite{nguyen2023climax} by introducing a Two-Branch Attention mechanism paired with Neural ODE \cite{chen2018neural} based residual updates inside each transformer block. The complete architecture diagram is shown in Figure \ref{fig:architecture}. Our attention uses a single shared query-key-value projection, but applies it through two computational paths: (i) a patch-wise self-attention branch and (ii) a derivative-based attention branch. Although both branches are initialized from the same projected tensors $(q,k,v)$, they achieve complementary roles because they apply different transformations and reshaping.

Given tokenized patch embeddings $x \in \mathbb{R}^{B \times N \times C}$, we compute a shared projection
$
[q,k,v] = W_{\mathrm{qkv}}x
$
and reshape into multi-head form
$
q,k,v \in \mathbb{R}^{B \times N_h \times N \times d_h},
$
where $C = N_h\cdot d_h$.

\paragraph{\textbf{Patch Attention (PA):}}
The first branch performs standard scaled dot-product attention over the patch tokens within each sample:
\begin{equation}
A^{\mathrm{pa}} = \mathrm{Softmax}\!\left(\frac{qk^\top}{\sqrt{d_h}}\right), 
\qquad
\mathrm{PA}(x) = A^{\mathrm{pa}}v,
\end{equation}
yielding $\mathrm{PA}(x) \in \mathbb{R}^{B \times N_h \times N \times d_h}$. This branch aggregates information across spatial patches, enabling each patch to attend to other regions of the same atmospheric state. It primarily captures spatial similarity and long-range dependencies, supporting global context and multi-scale pattern representation (e.g., coherent large-scale structures).

\paragraph{\textbf{Derivative-based attention (DA):}}
The second branch uses the same $(q,k,v)$ but reinterprets the computation by reshaping across the combined sample/head axis at each patch index:
\[
q_s,k_s,v_s \in \mathbb{R}^{N \times (BN_h) \times d_h}.
\]
Similarity logits are computed as
\begin{equation}
S = \frac{q_s k_s^\top}{\sqrt{d_h}},
\end{equation}
followed by a first-order finite difference of the logits along the re-indexed attention axis obtained by merging the batch and head dimensions (i.e., the $(B\!\cdot\!N_h)$ index), implemented by subtracting consecutive slices and padding:
\begin{equation}
\Delta S = S_{:,2:}-S_{:,:-1}, 
\qquad
\Delta S \leftarrow \mathrm{Pad}(\Delta S).
\end{equation}
Attention weights and the branch output are then
\begin{equation}
A^{\mathrm{da}} = \mathrm{Softmax}(\Delta S),
\qquad
\mathrm{DA}(x) = A^{\mathrm{da}} v_s.
\end{equation}
The result is reshaped back to $\mathbb{R}^{B \times N_h \times N \times d_h}$. Rather than using raw similarity scores, this branch operates on changes in similarity induced by the finite-difference operator. Consequently, it emphasizes relative variation in interaction strengths (i.e., where attention logits change sharply), providing a complementary change-sensitive signal to PA. This helps highlight transitions in patch relationships (e.g., boundaries or rapidly varying interaction patterns), even though the branch starts from the same shared $(q,k,v)$ projections.

\paragraph{\textbf{Fusion.}}
The two branch outputs are concatenated and projected back to the embedding dimension:
\begin{equation}
\mathrm{TA}(x) 
= 
W_o\!\left[\mathrm{PA}(x)\,\Vert\,\mathrm{DA}(x)\right]
\in \mathbb{R}^{B \times N \times C},
\end{equation}
matching the implementation where concatenation occurs along the per-head feature dimension, followed by an output projection.

\subsection{Neural ODE-Based Residual Updates:}
The fused attention output is updated via residual updates formulated as initial value problems and approximated via adaptive numerical integration implemented as Neural ODEs. For a hidden state $h \in \mathbb{R}^{B \times N \times C}$, we define the ODE dynamics as an initial value problem:
\begin{equation}
\frac{dz(t)}{dt} = f_{\theta}(z(t), t), 
\qquad z(0)=h,
\end{equation}

The input hidden state h is passed as the initial condition $z(0)$ to an ODE solver, which integrates the learned dynamics $f_\theta$ which is parameterized as a two-layer MLP with ReLU activation and hidden dimension matching the input dimension C. The solution is computed from $t\in[0,1]$ using \texttt{odeint} with an adaptive solver (default \texttt{dopri5}) and specified tolerances. The output $z(1)$ serves as the updated hidden state.

\begin{equation}
\mathrm{ODESolve}(h) = z(1).
\end{equation}
One ODE-Transformer block then applies two ODE-governed residual updates: one to the normalized attention pathway and one to the normalized MLP pathway:
\begin{equation}
h \leftarrow h + \mathrm{ODESolve}\!\left(\mathrm{LN}(\mathrm{TA}(h))\right),
\qquad
h \leftarrow h + \mathrm{ODESolve}\!\left(\mathrm{LN}(\mathrm{MLP}(h))\right).
\end{equation}
These continuous-depth updates replace discrete residual transformations, promoting smoother feature evolution through the continuous ODE formulation while retaining the standard transformer decomposition into attention and feed-forward components.

\subsection{Latitude Weighted Physics Informed Loss Function}
We use latitude weighted mean squared error to compute loss for the predicted variables.
\begin{equation}
    L_{lat-weight} = \frac{1}{V \times H \times W}\sum_{n=1}^N\sum_{i=1}^H\sum_{j=1}^W (L_i)(\hat{X}_{t+\Delta t}^{n,i,j}-X_{t+\Delta t}^{n,i,j})^2
\end{equation}
where $L_i$ accounts for Latitude weights:
\begin{equation*}
     L(i) = \frac{\cos(lat(i))}{\frac{1}{H}\sideset{}{_{i'=1}^{H}}\sum\cos(lat(i'))}
\end{equation*}
Additionally, we introduce two physics-motivated auxiliary penalty terms inspired by kinetic energy and temperature advection. Kinetic energy in atmospheric science refers to the energy associated with the motion of air masses. Thermodynamic balance equation describes the evolution of temperature in an air parcel due to processes like heat addition and pressure changes.

\begin{equation}
    \label{KEloss}
    L_{kinetic} = \left|\frac{1}{2}(u^2_{pred} + v^2_{pred}) - \frac{1}{2}(u^2_{true} + v^2_{true})\right|
\end{equation}
where u is the eastward wind component (m/s) and v is the northward wind component (m/s).

We compute a thermodynamic balance penalty by forming first-order finite-difference approximations of the temperature tendency and spatial gradients, and then minimizing the mean squared residual of an advection balance. 

\begin{equation}
L_{\mathrm{thermo}}
= \mathbb{E}\left[
\left(
\frac{\Delta T}{\Delta t}
+
u\frac{\Delta T}{\Delta x}
+
v\frac{\Delta T}{\Delta y}
\right)^2
\right],
\end{equation}

where $\frac{\Delta T}{\Delta t}$, $\frac{\Delta T}{\Delta x}$ and $\frac{\Delta T}{\Delta y}$ are first-order finite-difference approximations of the temperature tendency and spatial gradients computed on the native grid. The terms $\frac{\Delta T}{\Delta x}$ and $\frac{\Delta T}{\Delta y}$ approximate horizontal advection using the predicted wind components $(u,v)$.

\begin{equation}
    Loss_{physics} = L_{lat-weight} + \alpha \times L_{kinetic} + \beta \times L_{thermo}
\end{equation}

 where $\alpha $ and $\beta$ are the weight factors for physics based loss. The resulting the loss function is inspired by physical principles, however it does not enforce strict physical laws but instead guides the model toward outputs that are consistent with physical expectations. 

We compute finite differences on the native grid without explicit metric scaling; the penalty therefore encourages directional consistency rather than enforcing a dimensionally exact advection residual. Incorporating latitude-dependent grid spacing is a straightforward extension for future work."

We note that both kinetic energy and temperature evolve over time due to forcing, dissipation, and energy conversion. Our loss terms therefore do not enforce strict conservation laws. Rather, they act as physics-motivated auxiliary penalties on physically meaningful derived quantities, encouraging the model to produce forecasts whose wind-derived kinetic energy closely tracks the ground truth. Similarly, $L_{\mathrm{thermo}}$ penalizes the advection-equation residual evaluated on the predicted fields themselves, thereby promoting internal consistency between predicted winds and temperature gradients without requiring explicit modeling of diabatic source terms. Together, these terms provide a richer gradient signal than per-variable MSE alone and guide learning toward physically coherent outputs.
 
\section{Experiments and Results}
We train our model on ERA5 \cite{hersbach2018era5,hersbach2020era5} at the resolution of \(5.625^\circ\) (32 x 64 grid points) provided by WeatherBench \cite{rasp2020weatherbench}. We compare our approach with discrete transformer network (non-pretrained ClimaX \cite{nguyen2023climax}) and physics informed continuous model ClimODE \cite{verma2023climode}. We evaluate at \(5.625^\circ\) to enable controlled comparison under limited compute, whether the observed gains persist at higher resolutions remains an open question for future work and is addressed in section \ref{conclusion}.

\subsection{Training Details.} 
We consider weather forecasting as a continuous spatio temporal forecasting problem i.e a tensor of shape \(V \times H \times W \) at time \(t\) is fed to the model as input and it outputs a tensor of \(V' \times H' \times W' \) at future time step \(t + \triangle t\). The complete set of hyperparameters are given in Table \ref{hyper}.

\begin{table}[!h]
\caption{Training Hyperparameters}
\label{hyper}
\begin{center}
\begin{tabular}{llc}
\toprule
\multicolumn{1}{l}{\bf Hyperparameters}  &\multicolumn{1}{l}{\bf Meaning} &\multicolumn{1}{l}{\bf Value}\\ 
\midrule
p & Patch size & 2 \\
Heads & Number of continuous attention heads  & 16 \\
Depth & Number of Transformer layers & 4 \\
Dimension & Hidden dimensions & 1024\\
Dropout & Dropout rate &  0.1 \\
batch\_size & Batch Size &  12  \\
epochs & Training Epochs & 50 \\
norm & Normalization & z-score \\
lr & Learning Rate & 5e-5 \\
$\beta1$, $\beta2$ & Decay Rate of AdamW optimizer &  0.9, 0.999 \\
ES & Early stopping &  True \\
$\alpha$ & Kinetic Loss weight factor  &  0.1-0.5 \\
$\beta$ & Thermodynamic Loss weight factor  &  0.8 \\
\bottomrule
\end{tabular}
\end{center}
\end{table}

\subsection{WeatherBench}
We train PINN-Cast on hourly data with following set of variables:  Land Sea Mask (LSM), Orography, 10-meter U and V wind components (U10 and V10) and 2-meter temperature (T2m) in addition to 6 atmospheric variables: geopotential (Z), temperature (T), U and V wind components, specific humidity (Q) and relative humidity (R) at 7 pressure levels: 50 250 500 600 700 850 925. We use data from 1979-2015 for training, 2016 for validation and 2017-2018 for testing phase. We compare STC-ViT with ClimaX, and ClimODE on ERA5 dataset at \(5.625^\circ\) resolution provided by WeatherBench \cite{rasp2020weatherbench}. To ensure fairness, we retrained ClimaX from scratch without any pre-training.

\subsection{Evaluation Metrics.}
\label{eval}
We used Root Mean Square Error (RMSE) and Anomaly Correlation Coefficient (ACC) to evaluate our model predictions. The formula used for RMSE is:

\begin{equation}
    RMSE = \frac{1}{N} \sum\limits^N_{k=1}\sqrt{ \frac{1}{H \times W} \sum\limits^H_{i=1}\sum\limits^W_{j=1}L(i)(\hat{X}_{k,i,j}-X_{k,i,j})^2}
\end{equation}
where \(H \times W\) is the spatial resolution of the weather input and N is the number of total samples used for training or testing. L(i) is used to account for non-uniformity in grid cells.

\begin{equation}
    ACC = 
     \frac{
     \sum_{k,i,j}\hat{X'}_{k,i,j}X'_{k,i,j}
     }
     {
     \sqrt{\sum_{k,i,j} L(i) \hat{X'}^2_{k,i,j}\sum_{k,i,j} L(i) X'^2_{k,i,j}}
     }
\end{equation}
Where 
    \(\hat{X'} = \hat{X'} - C\) and \(X' = X' - C \) 
and C is the temporal mean of the entire test set \( C = \frac{1}{N}\sum_kX\)

\subsection{Results}
Our model outperforms both approaches at all lead times which shows that replacing regular attention with two-branch attention in ViT architecture derives improved feature extraction by mapping the changes occurring between successive time steps. Additionally, enforcing physical constraints in the model lead to improved prediction scores. The RMSE and ACC results are  shown in Table \ref{results-5.6}.

\begin{table}[H]
\caption{RMSE and ACC results trained on ERA5 at \(5.625^\circ\) resolution}
\label{results-5.6}
\begin{center}
\begin{adjustbox}{max width=\textwidth,center}
\begin{tabular}{lc ccc ccc}
\toprule
&  &\multicolumn{3}{c}{RMSE (Lower is better)} &\multicolumn{3}{c}{ ACC (Higher is better)} \\ \cmidrule(lr{.85em}){3-5} \cmidrule(lr{.85em}){6-8}

\textbf{Variable} &  \textbf{Lead Time (hrs.)} & Ours & ClimODE & ClimaX  & Ours & ClimODE & ClimaX 

\\ \midrule
\multirow{5}{*}{T2m (K)}  & 
6  & \textbf{0.87} & 1.20 & 2.02 & \textbf{0.98} & 0.97 & 0.92 \\ & 
12 & \textbf{1.04} & 1.44 & 2.26 & \textbf{0.97} & 0.96 & 0.90 \\& 
18 & \textbf{1.13} & 1.42 & 2.45 & \textbf{0.97} & 0.96 & 0.88 \\ & 
24 & \textbf{1.18} & 1.40 & 2.37 & \textbf{0.97} & 0.96 & 0.89 \\ & 
36 & \textbf{1.42} & 1.70 & 2.85 & \textbf{0.96} & 0.94 & 0.84
\\ \midrule
\multirow{5}{*}{T850 (K)}  & 
6  & \textbf{0.83} & 1.16 & 1.64 & \textbf{0.98} & 0.97 & 0.94\\ & 
12 & \textbf{0.99} & 1.32 & 1.77 & \textbf{0.97} & 0.96 & 0.93 \\& 
18 & \textbf{1.09} & 1.47 & 1.93 & \textbf{0.97} & 0.96 & 0.92 \\ & 
24 & \textbf{1.19} & 1.55 & 2.17 & \textbf{0.97} & 0.95 & 0.90\\ & 
36 & \textbf{1.44} & 1.75 & 2.49 & \textbf{0.95} & 0.94 & 0.86
 \\ \midrule
\multirow{5}{*}{u10 \((m\backslash s)\)}  & 
6  & \textbf{0.92} & 1.44 & 1.58 & \textbf{0.97} & 0.91 & 0.92 \\ & 
12 & \textbf{1.11} & 1.80 & 1.96 & \textbf{0.96} & 0.89 & 0.88 \\& 
18 &  \textbf{1.28} & 1.97 & 2.24 & \textbf{0.95} & 0.88 & 0.84 \\ & 
24 & \textbf{1.46} & 2.00 & 2.49 & \textbf{0.93} & 0.87 & 0.80\\ & 
36 &  \textbf{1.89} & 2.25 & 2.95 & \textbf{0.89} & 0.83 & 0.70
\\ \midrule
\multirow{5}{*}{v10 \((m\backslash s)\)}  & 
6  & \textbf{0.95} & 1.53 & 1.60 & \textbf{0.97} & 0.92 & 0.92 \\ & 
12 & \textbf{1.15} & 1.81 & 1.97 & \textbf{0.96} & 0.89 & 0.88 \\& 
18 & \textbf{1.32} & 1.95 & 2.26 & \textbf{0.94} & 0.88 & 0.83 \\ & 
24 & \textbf{1.50} & 2.02 & 2.48 & \textbf{0.93} & 0.86 & 0.80 \\ & 
36 &  \textbf{1.92} & 2.29 & 2.94 & \textbf{0.88} & 0.83 & 0.70

\\ \bottomrule
\end{tabular}
\end{adjustbox}
\end{center}
\end{table}

\subsection{Qualitative Results}
\begin{figure}[H]
    \centering
    \begin{subfigure}[t]{0.3\textwidth}
        \centering
        \includegraphics[width=\textwidth]{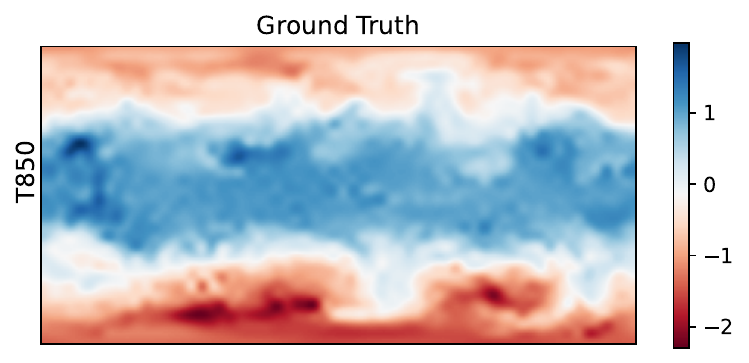}
    \end{subfigure}
    \begin{subfigure}[t]{0.3\textwidth}
        \centering
        \includegraphics[width=\textwidth]{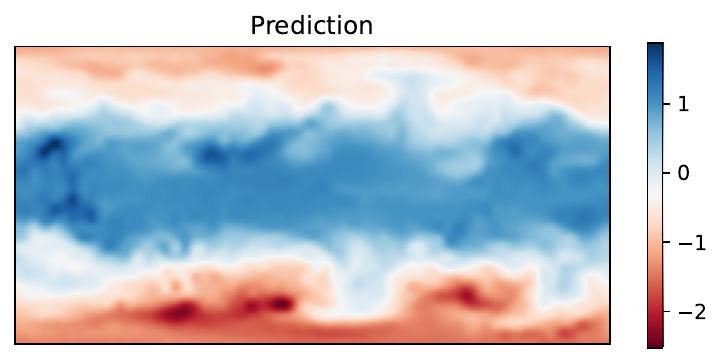}
    \end{subfigure}
    \begin{subfigure}[t]{0.3\textwidth}
        \centering
        \includegraphics[width=\textwidth]{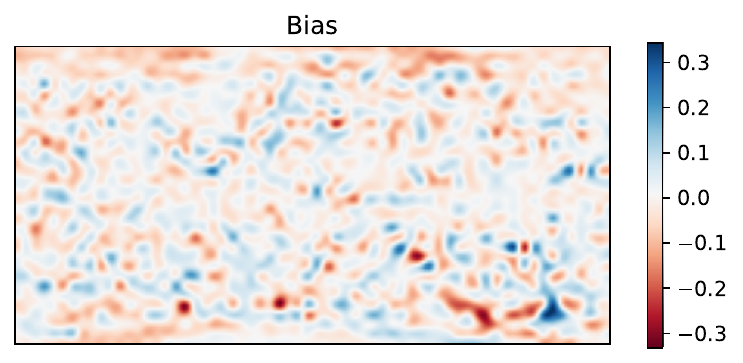}
    \end{subfigure}

    \begin{subfigure}[t]{0.3\textwidth}
        \centering
        \includegraphics[width=\textwidth]{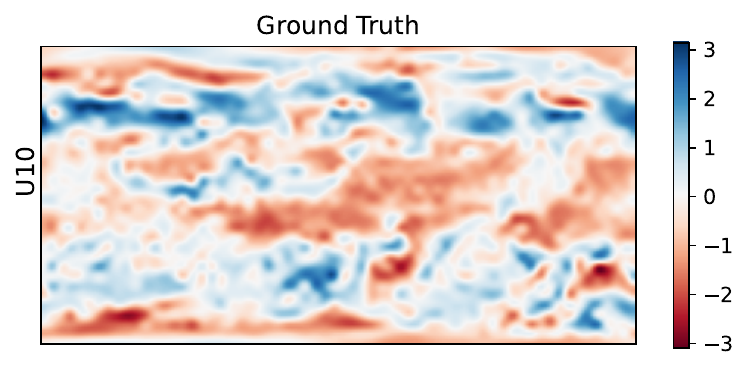}
    \end{subfigure}
    \begin{subfigure}[t]{0.3\textwidth}
        \centering
        \includegraphics[width=\textwidth]{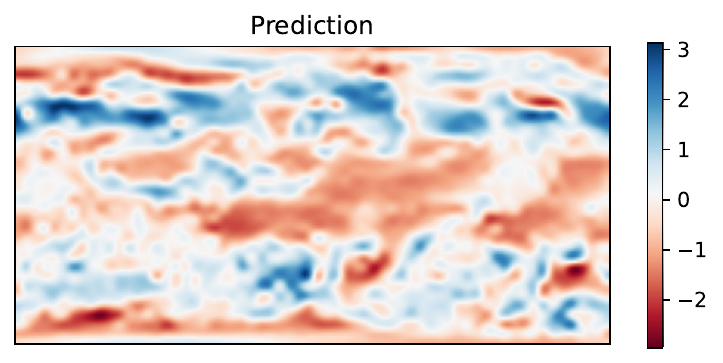}
    \end{subfigure}
    \begin{subfigure}[t]{0.3\textwidth}
        \centering
        \includegraphics[width=\textwidth]{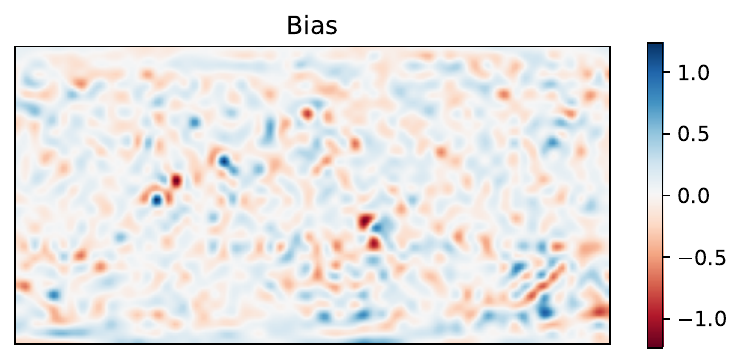}
    \end{subfigure}
    \begin{subfigure}[t]{0.3\textwidth}
        \centering
        \includegraphics[width=\textwidth]{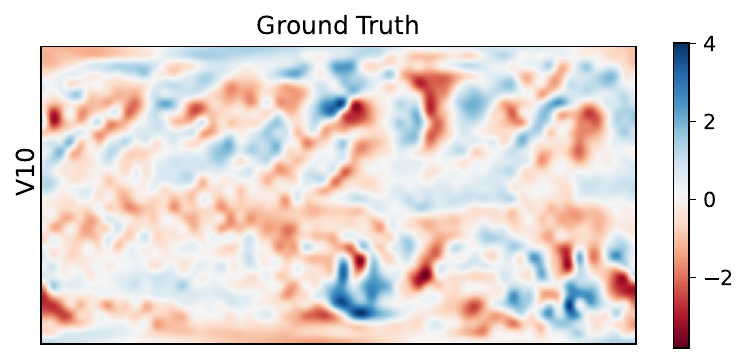}
    \end{subfigure}
    \begin{subfigure}[t]{0.3\textwidth}
        \centering
        \includegraphics[width=\textwidth]{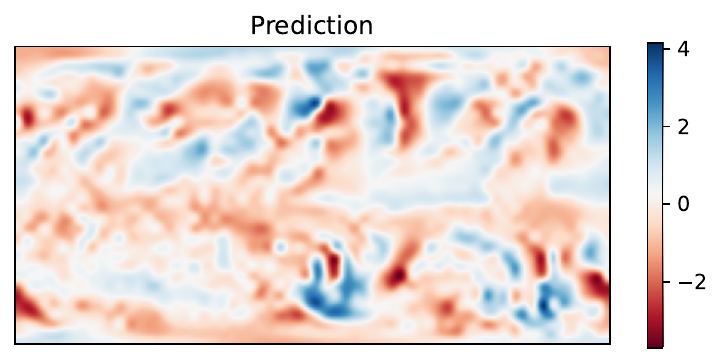}
    \end{subfigure}
    \begin{subfigure}[t]{0.3\textwidth}
        \centering
        \includegraphics[width=\textwidth]{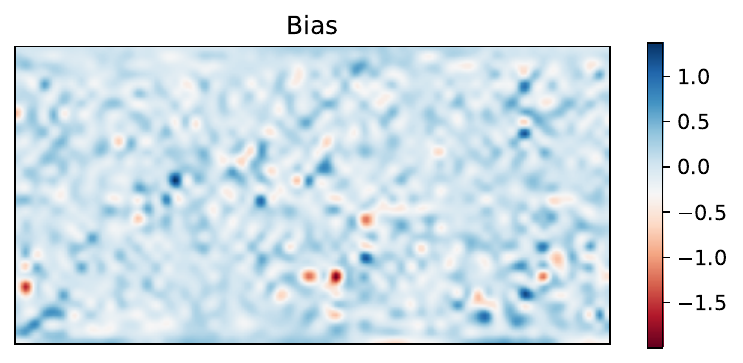}
    \end{subfigure}
    \caption{12hr forecast results} 
    \label{now}
\end{figure}

\begin{figure}[!h]
    \centering
    \begin{subfigure}[t]{0.3\textwidth}
        \centering
        \includegraphics[width=\textwidth]{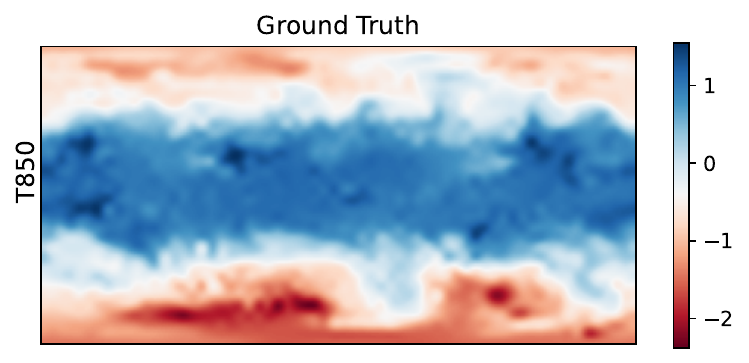}
    \end{subfigure}
    \begin{subfigure}[t]{0.3\textwidth}
        \centering
        \includegraphics[width=\textwidth]{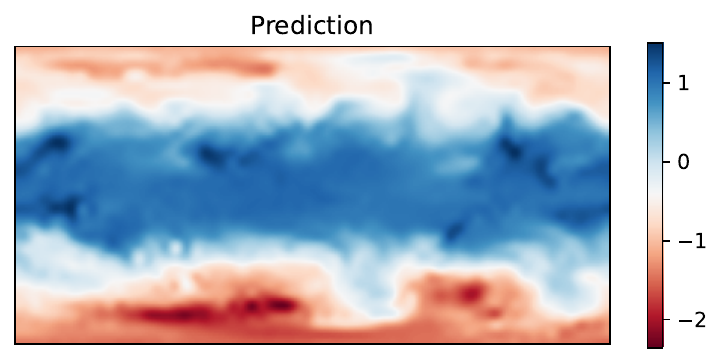}
    \end{subfigure}
    \begin{subfigure}[t]{0.3\textwidth}
        \centering
        \includegraphics[width=\textwidth]{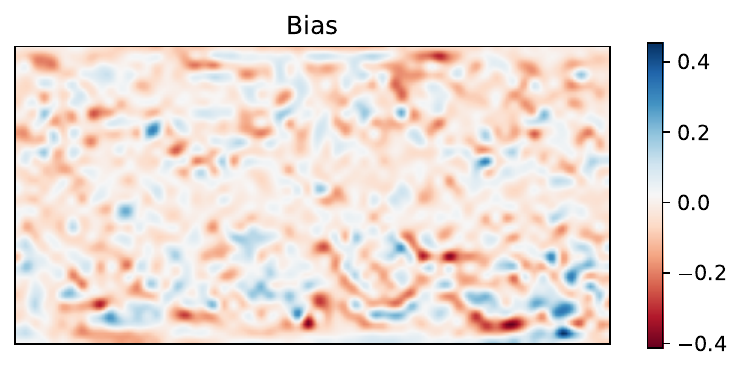}
    \end{subfigure}
    \begin{subfigure}[t]{0.3\textwidth}
        \centering
        \includegraphics[width=\textwidth]{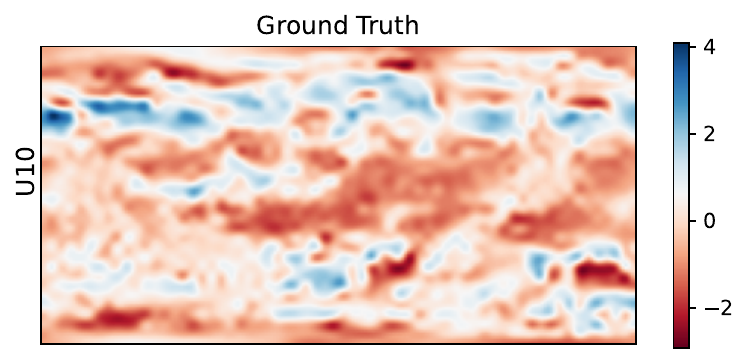}
    \end{subfigure}
    \begin{subfigure}[t]{0.3\textwidth}
        \centering
        \includegraphics[width=\textwidth]{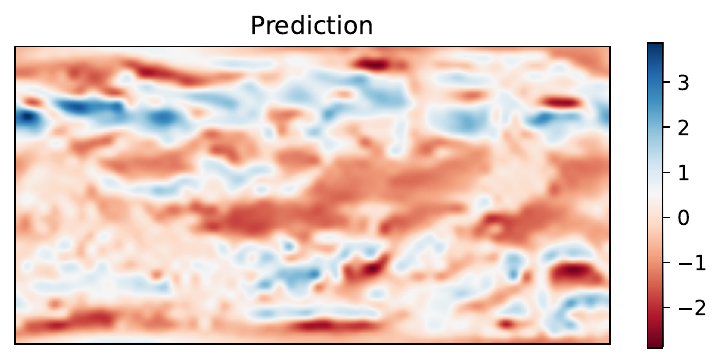}
    \end{subfigure}
    \begin{subfigure}[t]{0.3\textwidth}
        \centering
        \includegraphics[width=\textwidth]{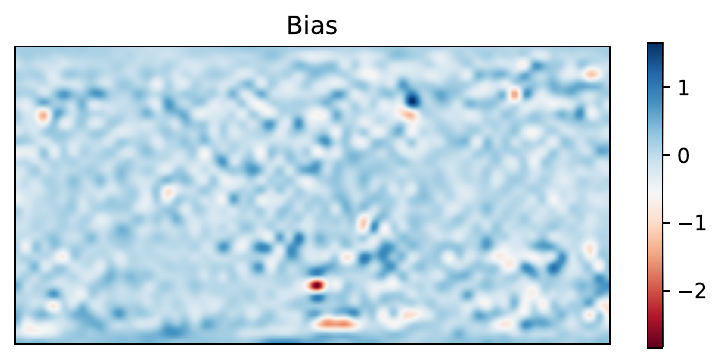}
    \end{subfigure}
    \begin{subfigure}[t]{0.3\textwidth}
        \centering
        \includegraphics[width=\textwidth]{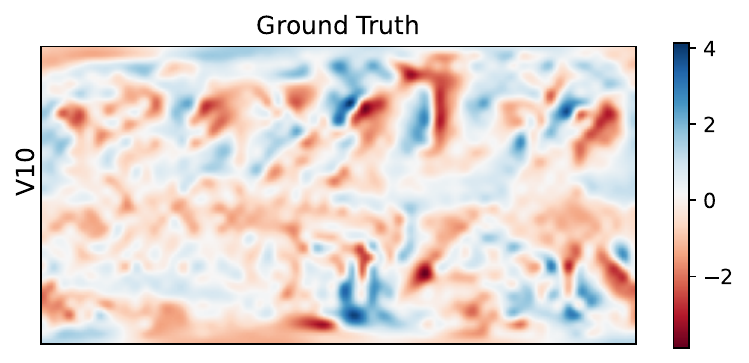}
    \end{subfigure}
    \begin{subfigure}[t]{0.3\textwidth}
        \centering
        \includegraphics[width=\textwidth]{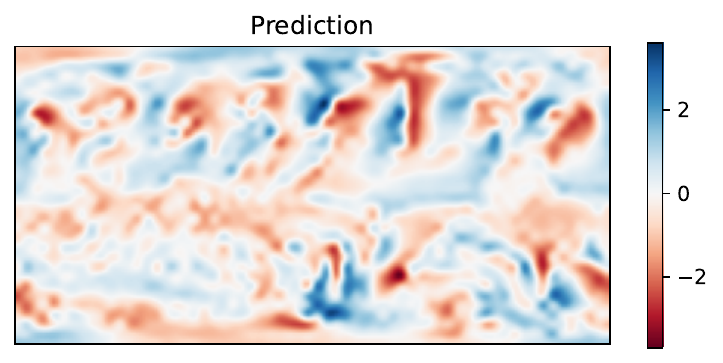}
    \end{subfigure}
    \begin{subfigure}[t]{0.3\textwidth}
        \centering
        \includegraphics[width=\textwidth]{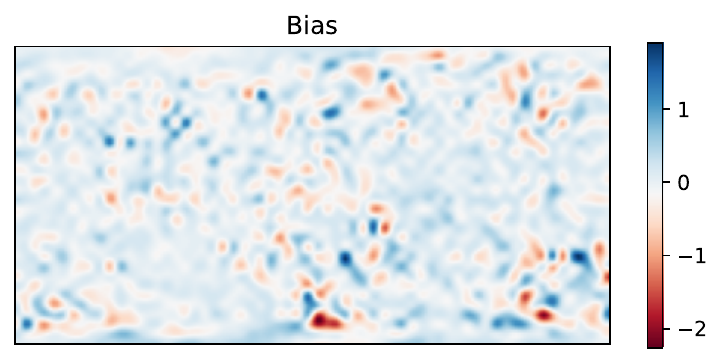}
    \end{subfigure}
    \caption{1 day forecast results} 
    \label{now}
\end{figure}

\subsection{Ablation Studies}
\label{experiment:ablation}
We perform ablation studies to isolate the contribution of each proposed component to overall forecast accuracy. The full PINN-Cast model combines both architectural modifications with the physics-informed loss, allowing us to assess whether the improvements stem from richer attention interactions, smoother continuous-depth dynamics, or physics-motivated supervision, and how these components interact when combined.

\paragraph{\textbf{Vanilla ViT.}} We train a standard Vision Transformer with single-branch self-attention, discrete residual updates, and latitude-weighted MSE loss, serving as our base architecture.

\paragraph{\textbf{Two-Branch Attention.}} We replace single-branch attention with our proposed two-branch attention module (PA + DA) to isolate the effect of the derivative-based interaction signal, while keeping discrete residual updates and MSE-only training.

\paragraph{\textbf{Neural ODE.}} We replace discrete residual updates with ODE-governed updates in the encoder blocks to evaluate the contribution of continuous-depth dynamics, while retaining single-branch attention and MSE-only loss.

\begin{figure}[!ht]
\begin{center}
\centerline{\includegraphics[width=\textwidth]{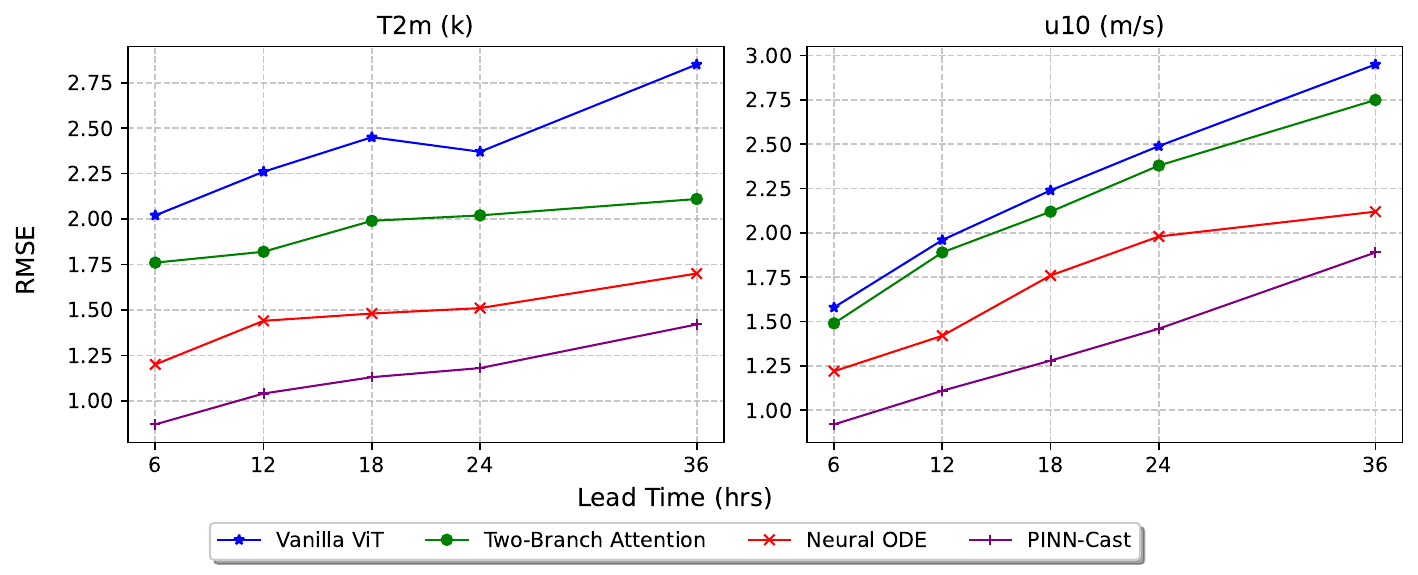}}
\caption{Ablation studies highlight the results for different components of PINN-Cast individually. Each row isolates a component of PINN-Cast. Each component yields improvement over the base ViT, while combining all three components (two-branch attention, ODE-governed updates, and physics-informed loss) achieves the best overall performance.}
\label{fig:ablation}
\end{center}
\end{figure}

\paragraph{\textbf{PINN-Cast (Full).}} Finally, we combine all three components: two-branch attention, ODE-governed residual updates, and physics-informed loss representing the complete proposed method. The addition of kinetic energy and advection consistency penalties on top of the architectural modifications further improves forecast quality, confirming that physics-motivated supervision provides complementary benefits beyond what the architectural changes alone achieve. The ablation results are shown in Figure\ref{fig:ablation}.

\subsection{Software and Hardware Requirements}
We use PyTorch \cite{paszke2019pytorch}, Pytorch Lightning \cite{falcon2019pytorch}, torchdiffeq \cite{chen2018neural}, and xarray \cite{hoyer2017xarray} to implement our model. We use 4 NVIDIA Tesla Volta V100-SXM2-32GB for the training at resolution of \(5.625^\circ\) with complete training time just under 2 days.

\section{Conclusion, Limitations, Future Work and Impact}
\label{conclusion}
\paragraph{Conclusion.}
We presented PINN-Cast, a transformer forecaster that integrates ODE-governed residual updates, a two-branch attention module with a derivative-based interaction signal, and physics-motivated auxiliary losses as soft constraints. Evaluation on WeatherBench at $5.625^\circ$ resolution against ClimaX and ClimODE demonstrates that each component contributes to improved short-term forecast accuracy, with the full model achieving the best performance across all variables and lead times.

\paragraph{Limitations and Future Work.}
Our evaluation is limited to a single coarse resolution ($5.625^\circ$) and a restricted set of baselines, whether the observed gains persist at higher resolutions and against pretrained or larger-scale forecasters remains an open question. The physics-informed penalties rely on unscaled finite differences and simplified advection proxies rather than dimensionally exact governing equations. Future work will extend to higher resolutions and additional variables, incorporate latitude-dependent grid scaling into the physics losses, broaden comparisons to recent forecasting models, and investigate how ODE dynamics and physics penalties shape learned representations.

\paragraph{Impact.}
This work provides a modular framework for integrating continuous-depth dynamics and physics-based regularization into transformer forecasters without hard-wiring constraints into the architecture. The approach offers a practical pathway toward data-driven weather models that balance predictive accuracy, physical plausibility, and computational efficiency across varying resource budgets.

\begin{credits}
\subsubsection{\ackname} This work was supported by SmartSAT Cooperative Research Center(project number P3-31s). We would like to acknowledge National Computational Infrastructure (NCI) (DOI: \url{10.26190/PMN5-7J50}), a high performance computing center for providing us with the GPU resources and weather data collection (WeatherBench and ERA5) which enabled us to perform this research.

\subsubsection{\discintname}
The authors have no competing interests to declare that are relevant to the content of this article.
\end{credits}
%
%
%
\bibliographystyle{splncs04}
\bibliography{mybibliography}

\end{document}